\def\year{2020}\relax
\documentclass[letterpaper]{article} % DO NOT CHANGE THIS
\usepackage{aaai20}  % DO NOT CHANGE THIS
\usepackage{times}  % DO NOT CHANGE THIS
\usepackage{helvet} % DO NOT CHANGE THIS
\usepackage{courier}  % DO NOT CHANGE THIS
\usepackage[hyphens]{url}  % DO NOT CHANGE THIS
\usepackage{graphicx} % DO NOT CHANGE THIS
\usepackage{graphicx}  % DO NOT CHANGE THIS
\usepackage[spacesep,definevectors]{easyvector}
\usepackage{amsmath}
\usepackage{mathtools}
\usepackage{xspace}
\usepackage{multirow}

\usepackage{algpseudocode}
\usepackage{algorithm}

\DeclareMathOperator*{\argmax}{argmax}
\newcommand{\Relu}{\mathsf{ReLU}} %% outer product

\newcommand{\largevec}[1]{\overrightarrow{#1}}
\newcommand\cev[1]{\overleftarrow{#1}}

\newcommand{\FGNET}{\text{FG-NET}}
\newcommand{\FGGCN}{\text{FGET-RR}}

\newcommand{\fixme}[1]{}

\begin{document}


%\title{Refined Fine Grained Named Entity Recognition using Graph Convolution Networks}
\title{Fine-Grained Named Entity Typing over Distantly Supervised Data Based on
  Refined Representations%
  \thanks{M.A.~ALi and W.~Wang are the co-corresponding authors.}
}

\author{
    Muhammad Asif Ali,\textsuperscript{\rm 1}
    Yifang Sun,\textsuperscript{\rm 1}
    Bing Li,\textsuperscript{\rm 1}
    Wei Wang,\textsuperscript{\rm 1,2}\\
    \textsuperscript{\rm 1}School of Computer Science and Engineering, UNSW, Australia\\
    \textsuperscript{\rm 2}College of Computer Science and Technology, DGUT, China\\
    \{muhammadasif.ali,bing.li\}@unsw.edu.au, \{yifangs,weiw\}@cse.unsw.edu.au\\
}

\maketitle

%% Abstract

\begin{abstract}
    Fine-Grained Named Entity Typing (FG-NET) is a key component in Natural Language Processing (NLP). It aims at classifying an entity mention into a wide range of entity types. Due to a large number of entity types, distant supervision is used to collect training data for this task, which noisily assigns type labels to entity mentions irrespective of the context. In order to alleviate the noisy labels, existing approaches on FG-NET analyze the entity mentions entirely independent of each other and assign type labels solely based on mention's sentence-specific context. This is inadequate for highly overlapping and/or noisy type labels as it hinders   information passing across sentence boundaries. For this, we propose an edge-weighted attentive graph convolution network that refines the noisy mention representations by attending over corpus-level contextual clues prior to the end classification. Experimental evaluation shows that the proposed model outperforms the existing research by a relative score of upto 10.2\% and 8.3\% for macro-f1 and micro-f1 respectively.
\end{abstract}

%% Introduction
%\input{1_Introduction.tex}

\section{Introduction}

%% 1    Introduction:
Named Entity Typing (NET) aims at classifying an entity mention to a set of entity types 
({e.g.,} person, location and organization) based on its context. It is one of the crucial 
components in NLP, as it helps in numerous down streaming applications, {e.g.,} information 
retrieval~\cite{DBLP:conf/wsdm/CarlsonBWHM10}, Knowledge Base Construction (KBC)~\cite{DBLP:conf/kdd/0001GHHLMSSZ14}, 
question answering~\cite{DBLP:conf/airs/LeeHOLHLKWJ06}, 
machine translation~\cite{DBLP:journals/corr/BritzGLL17}, {etc.} Fine-Grained Named Entity 
Typing (\FGNET) is an extension of traditional NET to a much wide range of entity types~\cite{DBLP:conf/emnlp/CorroAGW15,DBLP:conf/emnlp/RenHQHJH16}, typically over 
hundred types arranged in a hierarchical structure. It has shown promising results in 
different applications including KBC~\cite{DBLP:conf/kdd/0001GHHLMSSZ14}, relation extraction~\cite{mitchell2018never}, {etc.}

\begin{figure*}[ht]
    \centering
    \resizebox{0.944\linewidth}{!}{
    \includegraphics[]{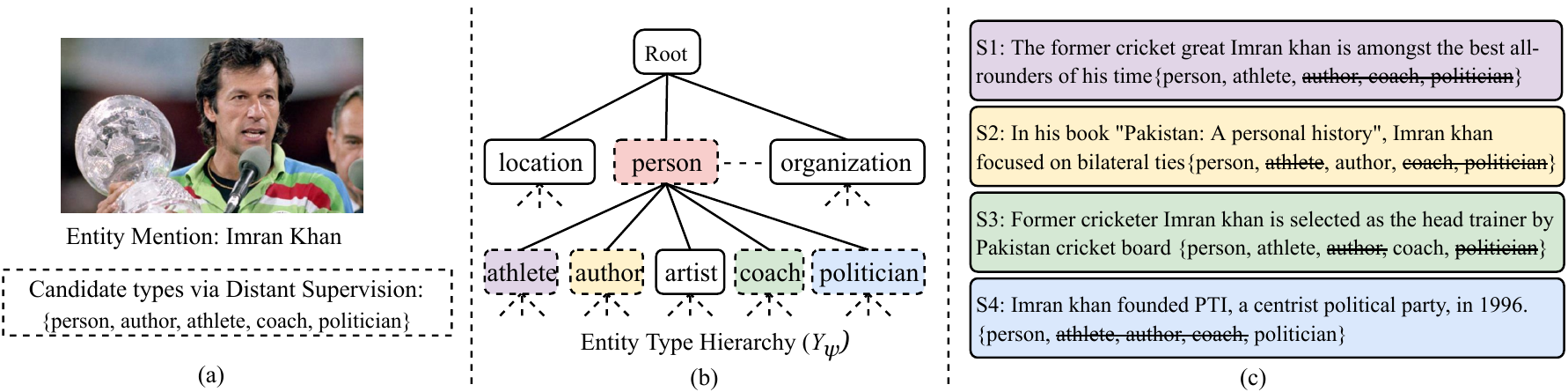}} %%! do not use suffix
    \caption{(a) Entity mention and candidate entity types acquired via distant supervision, (b) 
        Target Entity Type Hierarchy (c) Noisy training data with irrelevant entity types struck-through.}
    \label{fig:Distant_supervision}
\end{figure*}

In \FGNET{}, an entity mention is labeled with multiple overlapping entity types based on 
the context. For instance, in the sentence: \emph{``After having recorded his role, Trump 
    spent the whole day directing the movie."} 
\emph{Trump} can be annotated as both \emph{actor} and \emph{director} at the same time. 
Owing to a broad range of highly correlated entity types with small contextual differences 
(Section~\ref{all:analysis}), manual labeling is error-prone and time-consuming, thus 
distant supervision is widely used to automatically acquire the training data. Distant 
supervision follows a two-step approach, {i.e.,} detecting the entity mentions followed 
by assigning type labels to the mentions using existing knowledge bases. However, 
it assigns type labels irrespective of the mention's context, which results in high label 
noise~\cite{DBLP:conf/kdd/RenHQVJH16}. 
This phenomenon is illustrated in Figure~\ref{fig:Distant_supervision}, where, for the 
sentences denoted as: S1:S4, the entity mention \textit{``Imran Khan"} is labeled with all possible labels 
in the knowledge-base \emph{\{person, author, athlete, coach, politician\}}. Whereas, from the contextual perspective, in S1 the mention should be labeled as \emph{\{person, athlete\}}; in S2 it should be assigned labels \emph{\{person, author\}}, {etc}. This label noise propagates in model learning, which hinders the improvement in performance.

In an attempt to deal with the noisy training data, existing research on \FGNET{} relies on the following different approaches: 
{(i)} assume all labels to be correct ~\cite{DBLP:conf/aaai/LingW12,DBLP:conf/acl/YogatamaGL15}, 
which severely affects the model performance; 
{(ii)} apply different pruning heuristics to prune the noisy labels~\cite{DBLP:journals/corr/GillickLGKH14}, 
however, these heuristics drastically reduce the size of training data;
{(iii)} bifurcate the training data into two categories: clean and noisy, if the type labels 
correspond to the same type path or otherwise~\cite{DBLP:conf/emnlp/RenHQHJH16,DBLP:conf/eacl/AbhishekAA17}, 
they ignore the fact that the labels, even corresponding to the same type path, may be noisy.
For these approaches, it is hard to guarantee that underlying modeling 
assumptions will have a substantial impact on alleviating the label noise. 
In addition, these approaches model the entity mentions entirely independent 
of each other, which hinders effective propagation of label-specific contextual 
information across noisy entity mentions.

\begin{figure*}[ht!]
    \centering
    \resizebox{0.93\linewidth}{!}{
        \includegraphics[]{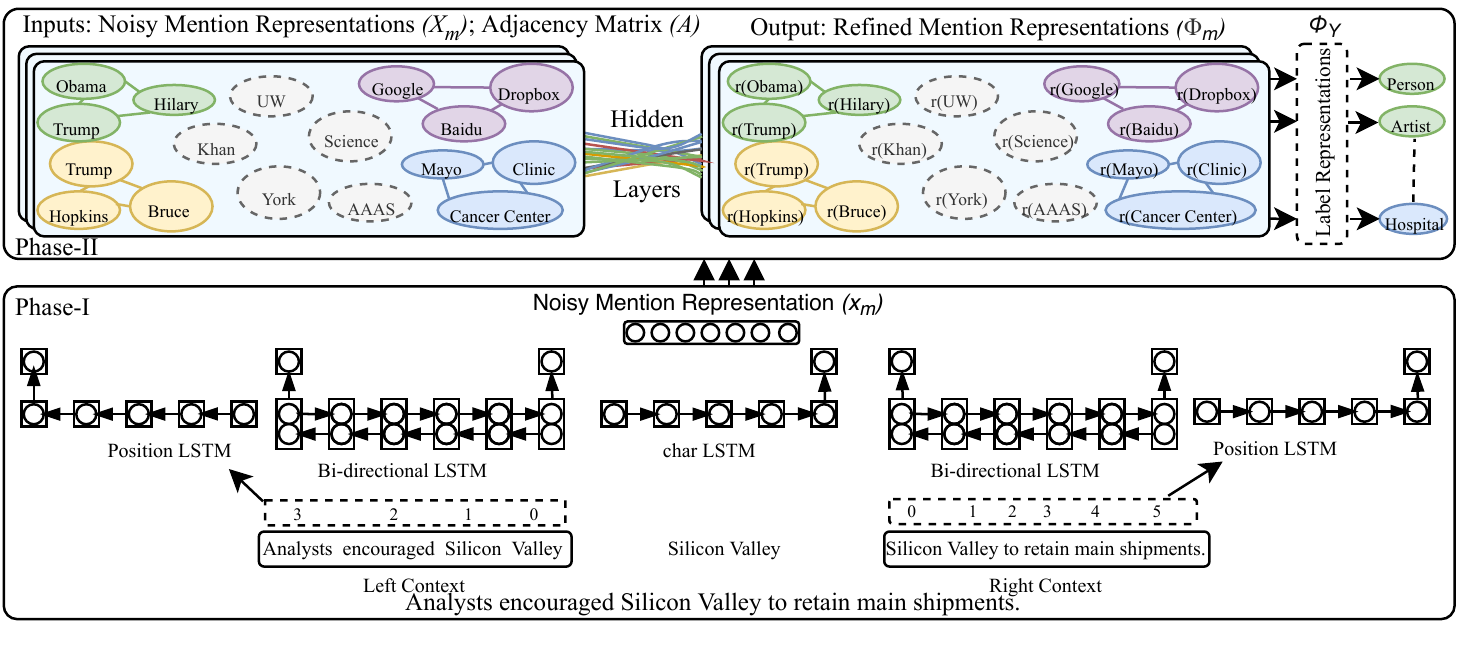}} %%! do not use suffix
    \caption{Proposed model for \FGNET{} (\FGGCN{}); Phase-I learns mention's representation based on local sentence-specific context; Phase-II refines the representations learnt in Phase-I by sharing corpus-level type-specific context}
    \label{fig:phase_all}
\end{figure*}

In order to address the challenges associated with the noisy training data, we introduce a novel 
approach that puts an equal emphasis on analyzing the entity mentions \textit{w.r.t} label-specific corpus-level 
context in addition to the sentence-specific context.
Specifically, we propose Fine-Grained named Entity Typing with Refined Representations (\FGGCN{}), 
shown in Figure~\ref{fig:phase_all}. \FGGCN{} initially uses mention's sentence-specific context 
to generate the noisy mention representation (Phase-I). Later, it uses corpus-level contextual 
clues to form a sparse graph that surrounds a subset of noisy mentions with a set of confident 
mentions having high contextual overlap. And, performs edge-weighted attentive graph convolutions 
to recompute/refine the representation of noisy mention as an aggregate of the confident neighboring 
mentions lying at multiple hops (Phase-II). Finally, the refined mention representation is embedded 
along with the type label representations for entity typing.

We argue that the proposed framework has following advantages: {(i)} it allows appropriate information 
sharing by efficient propagation of corpus-level contextual clues across noisy mentions; {(ii)} it 
analyzes the aggregated label-specific context, which is more refined compared with the noisy 
mention-specific context; {(iii)} it effectively correlates the local (sentence-level) and 
the global (corpus-level) context to refine mention's representation, required to perform the 
end-task in a robust way. We summarize the major contributions of this paper as follows:

\begin{itemize}
    \item We introduce \FGGCN{}, a novel approach for \FGNET{} that pays an equal importance on analyzing the entity mentions with respect to the corpus-level context in addition to the sentence-level context to perform entity typing in a performance-enhanced fashion.
    
    %\item We propose an edge-weighted attentive graph convolution network to refine noisy mention 
    %representations. To the best of our knowledge, this is the first work on \FGNET{} that, in contrast 
    %to existing models, refines representations learnt over distantly supervised training data.
    
    \item We propose an edge-weighted attentive graph convolution network to refine the noisy mention representations. To the best of our knowledge, this is the first work that, in contrast to the existing models that de-noise the data at model's input, refines the representations learnt over distantly supervised data.
    
    \item We demonstrate the effectiveness of the proposed model by comprehensive experimentation. \FGGCN{} outperforms the existing research by a margin of upto~10.2\% and ~8.3\% in terms of 
    macro-f1 and micro-f1 scores respectively.
    
\end{itemize}

%% Related_Work
%\input{2_Related_Work.tex}

\section{Related Work}
\label{related_work}

Earlier research on NET relies on assigning entity mentions to a small number of entity types, {i.e.,} person, location, organization, {etc.}~\cite{DBLP:conf/conll/SangM03}. In the recent decade, the traditional NET is extended to a wide range of fine-grained entity types~\cite{DBLP:conf/aaai/LingW12,DBLP:conf/acl/YosefBHSW13}. All the systems in \FGNET{} majorly focus on entity typing only, {i.e.,} they assume that the mention boundaries have been pre-identified.  
\citeauthor{DBLP:conf/acl/YogatamaGL15}, \shortcite{DBLP:conf/acl/YogatamaGL15} used embeddings to jointly embed entity mentions and the type information. \citeauthor{DBLP:journals/corr/GillickLGKH14}, \shortcite{DBLP:journals/corr/GillickLGKH14} proposed pruning heuristics to prune the noisy mentions. \citeauthor{DBLP:conf/emnlp/CorroAGW15}, \shortcite{DBLP:conf/emnlp/CorroAGW15} introduced the most fine-grained system so far, with types encompassing Word-Net Hierarchy~\cite{miller1998wordnet}.
\citeauthor{DBLP:conf/emnlp/RenHQHJH16}, \shortcite{DBLP:conf/emnlp/RenHQHJH16} introduced Automated Fine-grained named Entity Typing (AFET) using a set of hand-crafted features to represent mention, later jointly embed the feature vectors and the label vectors for classification. %\citeauthor{DBLP:conf/kdd/RenHQVJH16}, \shortcite{DBLP:conf/kdd/RenHQVJH16} introduced Label Noise Reduction (LNR) majorly focused on reducing the noise associated with the data acquired via distant supervision, later used it for \FGNET{}.

\citeauthor{DBLP:conf/akbc/ShimaokaSIR16}, \shortcite{DBLP:conf/akbc/ShimaokaSIR16} used an averaging encoder to encode the entity mention, bi-directional LSTM to encode the context, followed by attention to attend over label-specific context. \citeauthor{DBLP:conf/eacl/InuiRSS17}, \shortcite{DBLP:conf/eacl/InuiRSS17} extended~\cite{DBLP:conf/akbc/ShimaokaSIR16} by incorporating hand-crafted features along with attention. \citeauthor{DBLP:conf/eacl/AbhishekAA17}, \shortcite{DBLP:conf/eacl/AbhishekAA17} used end-to-end architecture to encode entity mention and its context. \citeauthor{DBLP:conf/naacl/XuB18}, \shortcite{DBLP:conf/naacl/XuB18} modified the \FGNET{} problem definition from multi-label classification to single-label classification problem with a hierarchical aware loss to handle noisy data.
\citeauthor{DBLP:conf/emnlp/XinZH0S18}, \shortcite{DBLP:conf/emnlp/XinZH0S18} proposed \FGNET{} based on language models that compute compatibility between the type labels and the context to eliminate inconsistent types.

Graph Convolution Networks (GCNs) have received considerable research attention in the recent past. 
GCNs extend the convolutions from regular grids to graph-structured data in spatial and/or spectral domain. 
They are widely been used in classification settings, {i.e.,} both 
semi-supervised~\cite{DBLP:conf/iclr/KipfW17}, and supervised~\cite{DBLP:conf/aaai/YaoM019}. 
While GCNs have successfully been used for image de-noising~\cite{DBLP:journals/corr/abs-1907-08448}, 
we are the first to effectively utilize it to refine the representations learnt over noisy text data. %for \FGNET.

%Major limitation of the existing research is the inability to cope with the noisy type labels in distant supervision.

%% The Proposed Model
%\input{3a_Proposed_Phase_1.tex}

\section{The Proposed Model}
\label{Proposed_model}

\subsection{Problem Definition}
% Definition

In this paper, we aim to build a multi-label, multi-class entity typing system that can use distantly supervised data to classify an entity mention into a set of fine-grained entity types based on the context. 
Specifically, we refine the representations learnt on the noisy data prior to entity typing. Similar to 
the existing research (used for comparative evaluation in Table~\ref{tab:Results}), we 
assume the availability of training data $D_{train}$ acquired via distant supervision and manually 
labeled test data $D_{test}$. 
Formally, the data set $D$ is a set of sentences/paragraphs for which the entity mentions $\{m_i\}_{i=1}^N$ 
(tokens corresponding to the entities), the context $\{c_i\}_{i=1}^N$ and the candidate type labels 
$\{y_i\}_{i=1}^N \in \{0,1\}^{Y}$ ($Y$-dimensional binary vector with $y_{i,t} = 1$ if $t^{th}$ 
type corresponds to the true label and zero otherwise) have been pre-identified. Here, the type 
labels correspond to type hierarchy in the knowledge base $\psi$ with the schema $Y_{\psi}$. 
We represent the data as a set of triples $D = \{(m_{i},c_{i},y_{i})\}_{i=1}^N$.
Following~\cite{DBLP:conf/emnlp/RenHQHJH16}, we bifurcate the training mentions $M_{train}$ 
into clean $M_{clean}$ and noisy $M_{noisy}$ depending upon if the mention's type path corresponds to a 
single path in $Y_{\psi}$ or otherwise. For example, considering the type-path in 
Figure~\ref{fig:Distant_supervision} (b), a mention with labels \emph{\{person, athlete, author\}} 
will be considered as a noisy, whereas, a mention with labels \emph{\{person, artist\}} will be considered as clean.

\subsection{Overview}
Our proposed model (shown in Figure~\ref{fig:phase_all}) consists of two phases: in Phase-I, we learn local context-dependent noisy mention representations using LSTM 
networks~\cite{DBLP:journals/neco/HochreiterS97}. In Phase-II, we form a sparse graph that 
takes the representations learnt in Phase-I as input and perform edge-weighted attentive 
graph convolutions to refine these representations. Finally, we embed the refined mention 
representations along with the label representations for \FGNET{}. 

We argue that the proposed two-phase design has the following advantages: {(i)} it allows us to quantify the contribution of each phase, as it provides the maximal flexibility to correlate and/or analyze these phases independently, {(ii)} it enables effective propagation of corpus-level contextual information that facilitates refinement of noisy mention representations.

\subsection{Phase-I (Noisy Mention Representations)}
\label{phase_1}
Phase-I follows a standard approach with multiple LSTM networks to encode sequential text data. 
We use $\largevec{x}$ and $\cev{x}$ to represent the left-to-right and the right-to-left LSTM encodings.
The components of Phase-I are explained as follows: 

\paragraph{Mention Encoder:}
To encode the morphological structure of entity mentions, we first decompose the
mention into character sequence. Later, use a standard LSTM network to encode
the character sequence. We use $\phi_{men} = [\largevec{men_{char}}] \in
\mathbf{R}^{d}$ to represent the encoded mention.

\paragraph{Context Encoder:}
In order to encode the context, we use bidirectional LSTMs to encode the tokens corresponding to the left 
and the right context of the entity mention, as shown in Figure~\ref{fig:phase_all}. Note 
that for each bi-directional LSTM, we feed mention tokens along with the context to get the context encoding. 
The motivation is to analyze the context in relation with the entity mention.
We use $\phi_{left}$ = $[\cev{c_{left}} ; \largevec{c_{left}}] \in \mathbf{R}^{c}$  , and $\phi_{right}$ = $[\cev{c_{right}} ; \largevec{c_{right}}] \in \mathbf{R}^{c}$ to represent bi-directional encoding of the left and the right context respectively.
%surrounding the entity mention
\paragraph{Position Encoder:}

The position feature is used to encode the relative distance between the mention and the contextual words 
using LSTM network. Previously, this feature has shown good performance in relation classification~\cite{DBLP:conf/coling/ZengLLZZ14}. 
We use $\phi_{lpos} = [\cev{left_{pos}}] \in \mathbf{R}^{p}$ and $\phi_{rpos} =  [\largevec{right_{pos}}] \in \mathbf{R}^{p}$ 
to encode the relative positions of the left and the right contextual tokens.

\paragraph{Mention Representation:}
Finally, we concatenate all the mention-specific encodings to get the noisy mention representation: $x_m \in \mathbf{R}^{f}$, 
where $ f = {d} + {2*c} + {2*p}$

\begin{equation}
\label{concat_eq}
    x_{m} = [\phi_{lpos} ; \phi_{left} ; \phi_{men} ; \phi_{right} ; \phi_{rpos}]    
\end{equation}

%% The Proposed Model
%\input{3b_Proposed_Phase_2.tex}

\subsection{Phase-II (Refining Mention Representations)}
In order to refine the noisy mention representations learnt on distantly supervised data (Phase-I), we propose an edge-weighted attentive Graph Convolution Network (GCN). GCN extends convolution from regular structured grids to arbitrary graphs. We exploit the fact that for a given graph node, the GCN uses the information contained in the neighboring nodes to come up with a new representation of the node. For this, we construct an undirected graph with nodes as entity mentions and enforce the mentions with high contextual overlap to be adjacent to each other by forming edges. Formally, let $G=(V, E)$ be a graph with $|V|=n$ nodes (entity mentions); $|E|$ edges; we use $A$ to denote its symmetric adjacency matrix. The construction of $G$ is outlined in Algorithm~\ref{alg:loop} and explained as follows:

\paragraph{Graph Construction:}
Firstly, we learn $1024d$  deep contextualized ELMO embeddings~\cite{DBLP:conf/naacl/PetersNIGCLZ18} for all the sentences in data set $D$. We average out the embedding vectors corresponding to the 
mention tokens to acquire context-dependent mention embeddings $ELMO_{men}$. 
Later, for the training data $D_{train}$, we compute pivot vectors, {i.e.,} $\{Pivot_{y} \}_{y=1}^{Y}$, as representatives for each entity type $y \in Y$, by averaging the mention embeddings corresponding to the type $y$ ($ELMO_{men_{y}}$). We use these pivot vectors to capture confident mention candidates for each entity type $\{Candidates\}_{y=1}^{Y}$, {i.e.,} the mentions with high contextual overlap having $cos(ELMO_{men},\{Pivot_{y}\}_{y=1}^{Y}) \geq thr$, as illustrated in lines (7-13) of Algorithm~\ref{alg:loop}. We observed that a reasonably high value for the threshold $thr$ offers the following benefits: (i) avoids computational overhead, (ii) captures only the most confident mention candidates. Finally, for the candidate mentions corresponding to each entity type $\{Candidates\}_{y=1}^{Y}$, we form pairwise edges to construct the graph $G$, with adjacency matrix $A$ (line 14-16).

\paragraph{Attentive Aggregation:}
Depending upon the value of $thr$, the graph $G$ surrounds a subset of nodes (noisy mentions), with a set of confident mentions having high type-specific contextual overlap, by forming edges. Later, for noisy mention representations, we aggregate the information contained in the neighbors to come up with the refined mention representations. Specifically, unlike the existing work~\cite{DBLP:conf/iclr/KipfW17}, we propose an edge-weighted attentive graph convolution network that uses the following layer-wise propagation rule:
\begin{equation}
\label{eq:FGNET-RR}
L^{(1)} = \rho(\widetilde{\eta_{ij} \odot A} L^{(0)} W_0)
\end{equation}
where $\eta_{ij}$ is the attention term computed via pairwise similarity of the context-dependent mention embeddings, {i.e.,} $cos(ELMO_{men^{i}}, ELMO_{men^{j}})$ $\forall (i,j) \in V$; ${\eta_{ij} \odot A}$ is the Hadamard product of attention weights and the adjacency matrix; $\widetilde{\eta_{ij} \odot A} = \tilde{D}^{-1/2} ((\eta_{ij} \odot A) + I) \tilde{D}^{-1/2}$ is 
the normalized symmetric matrix; $\tilde{D}$ is the degree matrix of $(\eta_{ij} \odot A)$; $L^{(0)}$ is input from the previous layer, in our case:
\begin{algorithm}[t]
	\caption{Graph Construction}
	\label{alg:loop}
	\textbf{Input:} {Embeddings $(ELMO_{men})$; $D=D_{train}+D_{test}$}\\
	\textbf{Output:} {Graph: $G$}
	\begin{algorithmic}[1]
		\Statex
		\State {$\{Pivot_{y}\}_{y=1}^{Y}$ $\gets$ {$\mathbf{0}$}; $G \gets \emptyset$ }
		\For{$men \gets 1$ to $D_{train}$}
		\For{$y \in men_{labels}$}
		\State {$Pivot_{y}$ $\gets$ {$Pivot_{y} + ELMO_{men_{y}}$}}
		\EndFor
		\EndFor
		\State{$\{Candidates\}_{y=1}^{Y}$ $\gets \emptyset$ }
		\For{$men \gets 1$ to $D$}
		\State{$y^{*} = \argmax_{y \in Y} \cos{(ELMO_{men},Pivot_{y})}$}
		\If{$\cos{(ELMO_{men},Pivot_{y^{*}})} \geq thr$}
		\State{$Candidates_{y^{*}} \gets Candidates_{y^{*}} \cup men$}
		\EndIf        
		\EndFor
		\For{$y \gets 1$ to $Y$}
		\State{$G$ $\gets$ $G \cup \{edge(v_{1},v_{2}) \in Candidates_{y}\}$}
		\EndFor
		\State \Return {$G$}
		
	\end{algorithmic}
\end{algorithm}    
$L^{(0)} = X_m \in \mathbf{R}^{N\times{f}}$ is the matrix corresponding to the noisy mentions' 
representations from Equation~\eqref{concat_eq}, $\rho$ is the activation function and $W_0$ is the 
matrix of learn-able parameters. Note, by adding identity matrix $I$ to $(\eta_{ij} \odot A)$, the 
model assumes that every node $v \in V$ is connected 
to itself, {i.e,} $(v,v) \in E$. We observed, that for our problem, this simple symmetrically 
normalized edge-weighted attentive formulation outperforms the attention formulation of~\cite{DBLP:journals/corr/BahdanauCB14}. We can 
accumulate information from the higher-order neighborhood by stacking multiple layers:

\begin{equation}
    L^{(i+1)} = \rho(\widetilde{\eta_{ij} \odot A} L^{(i)} W_{i})
\end{equation}

where $i$ corresponds to the layer no., with $L^{(0)} = X_m$. For our model, we use a two-layered network to learn the refined mention representations ${\Phi_{m}} \in \mathbf{R}^{N \times {k}}$ as follows:

\begin{equation}
    \Phi_m = \widetilde{\eta_{ij} \odot A}(ReLU(\widetilde{\eta_{ij} \odot A}X_m W_0)) W_1
\end{equation}

\subsection{The Complete Model}
Let $\phi_m \in \mathbf{R}^{k}$ be a refined mention representation and $\{\phi_y\}_{y=1}^{Y} \in \mathbf{R}^{k}$ be the type label representations. For classification, we embed these representations in the same space. For this, we learn a function $f(\phi_{m},\phi_{y}) = \phi_{m}^{T}.\phi_{y}+bias_{y}$ that incorporates label bias $bias_{y}$ in addition to the label and the refined mention representations. We extend loss functions from our previous work~\cite{DBLP:conf/aaai/AliSZWZ19} to separately model the clean and the noisy entity mentions, as explained below:

\paragraph{Loss Function for clean data:}
\label{sec:loss-funct-clean}
In order to model the clean entity mentions $M_{clean}$, we use a margin-based loss to 
embed the mention representation close to its true label representations, while at the 
same time pushing it away from the false type labels. The loss function for modeling 
the clean entity mentions is shown in Equation~\eqref{eq:clean}.

\begin{equation}
  \label{eq:clean}
\begin{aligned}
  L_{clean} &= \sum_{y \in T_y}  \Relu(1 - f(\phi_{m},\phi_{y})) \\
    &+ \sum_{y^{'} \in T'_y} \Relu(1 + f(\phi_{m},\phi_{y^{'}}))
\end{aligned} 
\end{equation}

where $T_{y}$ represents the true labels and $T_{y^{'}}$ represents the false labels in $Y_{\psi}$.
%$\phi_{m}$ corresponds to the refined mention representation, $\phi_{y}$ corresponds to type label representation, 
\paragraph{Loss Function for noisy data:}
\label{sec:loss-funct-noisy}
In order to model the noisy entity mentions $M_{noisy}$, we use a variant of the loss 
function in Equation~\ref{eq:clean} to focus on the most relevant label among noisy 
type labels. The loss function for modeling the noisy entity mentions is illustrated in Equation~\eqref{eq:noisy}.

\begin{equation}
  \label{eq:noisy}
\begin{aligned}
  L_{noisy} &=  \Relu(1 - f(\phi_{m},\phi_{y^{*}})) \\
    &+ \sum_{y^{'} \in T'_y} \Relu(1 + f(\phi_{m},\phi_{y^{'}}))\\
    & y^{*} = \argmax_{y \in T_y} f(\phi_{m},\phi_{y})
\end{aligned} 
\end{equation}

where $y^{*}$ corresponds to the most relevant label among the set of noisy labels, $T_{y}$ 
represents the set of noisy labels and $T_{y^{'}}$ represents the false labels  in $Y_{\psi}$.

%$\phi_{m}$ corresponds to the refined mention representation, $\phi_{y}$ corresponds to type label representation, 
Finally, we minimize $L_{noisy}$ + $L_{clean}$ as the loss function of \FGGCN{}.

\paragraph{Model Training and Inference:}
Owing to the adjacency matrix $A$ involved in Phase-II, our current implementation 
trains two phases iteratively. Specifically, we repeat the following process till 
convergence: {(i)} perform mini-batch Stochastic Gradient Descent (SGD) in Phase-I, 
{(ii)} concatenate the noisy representations learnt in Phase-I (i.e., $X_{m}$) and 
perform gradient descent in Phase-II. We leave an appropriate formulation of SGD for Phase-II as 
future work. For inference, we use mention's refined representation $\phi_{m}$ and carry-out a 
top-down search in the type-hierarchy, i.e., we recursively select the type $y$ that yields the 
best score $f(\phi_{m},\phi_{y})$ until we hit a leaf node or the score falls below a threshold of zero.

%%% Experimentation
%\input{4_Experiments.tex}

\section{Experiments}
\label{experimentions}
\subsection{Dataset}

For evaluation, we use publicly available data sets provided by \cite{DBLP:conf/emnlp/RenHQHJH16}. 
Table~\ref{tab:dataset_FG_GCN} shows the statistics of these data sets. A detailed description is as follows:

\paragraph{Wiki/Figer:} Its training data consists of Wikipedia sentences automatically labeled via 
distant supervision by mapping the entity mentions to Freebase types~\cite{DBLP:conf/sigmod/BollackerEPST08}. 
The testing data consists of news reports manually labeled by ~\cite{DBLP:conf/aaai/LingW12}. 

\paragraph{OntoNotes:} It consists of sentences from newswire documents contained in OntoNotes corpus~\cite{weischedel2011ontonotes} mapped to Freebase types via DBpedia Spotlight~\cite{DBLP:conf/i-semantics/DaiberJHM13}. The testing data is manually annotated by~\cite{DBLP:journals/corr/GillickLGKH14}.

\paragraph{BBN:} It consists of sentences from the Wall Street Journal annotated by~\cite{weischedel2005bbn}. The training data is annotated using DBpedia Spotlight.%~\cite{DBLP:conf/i-semantics/DaiberJHM13}. 

\begin{table*}[htbp]
    \centering
    \resizebox{1.75\columnwidth}{!}{%
        \begin{tabular}{l|lll|lll|lll}
            \hline
            & \multicolumn{3}{c}{Wiki} & \multicolumn{3}{c}{OntoNotes} & \multicolumn{3}{c}{BBN}  \\
            \hline
            \hline
            & strict & mac-F1 & mic-F1 & strict   & mac-F1   & mic-F1  & strict & mac-F1 & mic-F1 \\
            \hline
            \hline
            \textbf{FIGER}~\cite{DBLP:conf/aaai/LingW12}     & 0.474  & 0.692  & 0.655  & 0.369    & 0.578    & 0.516   & 0.467  & 0.672  & 0.612  \\
            \textbf{HYENA}~\cite{DBLP:conf/acl/YosefBHSW13}    & 0.288  & 0.528  & 0.506  & 0.249    & 0.497    & 0.446   & 0.523  & 0.576  & 0.587  \\
            \textbf{AFET-NoCo}~\cite{DBLP:conf/emnlp/RenHQHJH16} & 0.526  & 0.693  & 0.654  & 0.486    & 0.652    & 0.594   & 0.655  & 0.711  & 0.716  \\
            \textbf{AFET-NoPa}~\cite{DBLP:conf/emnlp/RenHQHJH16} & 0.513  & 0.675  & 0.642  & 0.463    & 0.637    & 0.591   & 0.669  & 0.715  & 0.724  \\
            \textbf{AFET-CoH}~\cite{DBLP:conf/emnlp/RenHQHJH16}  & 0.433  & 0.583  & 0.551  & 0.521    & 0.680     & 0.609   & 0.657  & 0.703  & 0.712  \\
            \textbf{AFET}~\cite{DBLP:conf/emnlp/RenHQHJH16}  & 0.533  & 0.693  & 0.664  & \underline{0.551}    & 0.711    & 0.647   & \underline{0.670}  & 0.727  & 0.735  \\
            \textbf{Attentive}~\cite{DBLP:conf/akbc/ShimaokaSIR16} & 0.581  & 0.780   & 0.744  & 0.473    & 0.655    & 0.586   & 0.484  & 0.732  & 0.724  \\
            \textbf{FNET-AllC}~\cite{DBLP:conf/eacl/AbhishekAA17} & \underline{0.662}  & 0.805  & 0.770  & 0.514    & 0.672    & 0.626   & 0.655  & 0.736  & 0.752  \\
            \textbf{FNET-NoM}~\cite{DBLP:conf/eacl/AbhishekAA17}  & 0.646  & 0.808  & 0.768  & 0.521    & 0.683    & 0.626   & 0.615  & 0.742  & 0.755  \\
            \textbf{FNET}~\cite{DBLP:conf/eacl/AbhishekAA17}    & 0.658  & \underline{0.812}  & \underline{0.774}    & 0.522    & 0.685   & 0.633  & 0.604   & 0.741  & 0.757  \\
            \textbf{NFGEC+LME}~\cite{DBLP:conf/emnlp/XinZH0S18} & 0.629  & 0.806  & 0.770    & 0.529    & \underline{0.724}   & \underline{0.652}  & 0.607     &  \underline{0.743}   & \underline{0.760}  \\
            \hline
            \hline
            %\textbf{\FGGCN{} (Phase I)} &   0.661     &    0.807   &    0.767    &   0.531   &  0.694   &  0.638   & 0.616  & 0.755  & 0.765  \\
            
            \textbf{\FGGCN{} Phase I-II (Glove + Context Encoders)}  &  0.674  &  0.817 &  0.777  &  0.567  &  0.737  &  0.680     & \textbf{0.740}  & 0.811  & 0.817\\
            
            \textbf{\FGGCN{} Phase I-II (Contextualized Embeddings)}  &  \textbf{0.710}  &  \textbf{0.847} &  \textbf{0.805}  &  \textbf{0.577}  &  \textbf{0.743}  &  \textbf{0.685}     & 0.703  & \textbf{0.819}  & \textbf{0.823}\\
            \hline
    \end{tabular}}
    \caption{\FGGCN{} performance comparison against baseline models}
    \label{tab:Results}
\end{table*}

\begin{table}[htbp]
    \centering
    \resizebox{0.80\columnwidth}{!}{
    \begin{tabular}{l | rrr}
        \hline
        Dataset & Wiki & OntoNotes & BBN \\
        \hline
        Training Mentions & 2.6 M & 220398 & 86078 \\
        Testing Mentions & 563 & 9603 & 13187 \\
        \% clean mentions (training) & 64.58  & 72.61 & 75.92 \\
        \% clean mentions (testing) & 88.28 & 94.0 & 100 \\
        Entity Types & 128 & 89 & 47 \\
        Max hierarchy depth & 2 & 3 & 2 \\
        \hline
    \end{tabular}}
    \caption{Fine-Grained Named Entity Typing data sets}
    \label{tab:dataset_FG_GCN}
\end{table}

\subsection{Experimental Settings:}
\label{experimental_settings}

In order to come up with a unanimous platform for comparative evaluation,  we use the 
priorly defined data split by the existing models to training, test and dev sets. 
The training data is used for model training ({i.e.,} learning noisy representations in Phase-I 
and refinement in Phase-II). The dev set is used for parameter tuning and the 
model performance is reported on the test set. All the experiments are performed on 
Intel Xenon Xeon(R) CPU E5-2640 (v4) with 256 GB main memory and Nvidia Titan V GPU.

\paragraph{Hyperparameters:}

We separately analyze the performance of \FGGCN{} using $300d$ Glove 
embeddings~\cite{DBLP:conf/emnlp/PenningtonSM14} and $1024d$ deep contextualized ELMO embeddings~\cite{DBLP:conf/naacl/PetersNIGCLZ18}. Character, position and label embeddings are 
randomly initialized. For position and bi-directional context encoders, the hidden layer size of 
LSTM is set to $100d$. For mention encoder the hidden layer size is $200d$. Maximum sequence length 
is set to 100. For Phase-II, we use graphs with 1.6M, 0.6M, 5.4M edges for Wiki, 
Ontonotes, and BBN respectively. For model training, we use Adam optimizer~\cite{DBLP:journals/corr/KingmaB14} 
with learning rate (0.0008-0.001).

\subsection{Baseline Models / Model Comparison}
\label{baselines__}

We compare \FGGCN{} with the existing state-of-the-art research on~\FGNET{}, namely: {(i)} \textbf{FIGER}~\cite{DBLP:conf/aaai/LingW12}, {(ii)} \textbf{HYENA}~\cite{DBLP:conf/acl/YosefBHSW13}, {(iii)} \textbf{AFET}~\cite{DBLP:conf/emnlp/RenHQHJH16} and its variants \textbf{AFET-NoCo}, \textbf{AFET-NoPa}, \textbf{AFET-CoH},
{(iv)} \textbf{Attentive}~\cite{DBLP:conf/akbc/ShimaokaSIR16}, {(v)} \textbf{FNET}~\cite{DBLP:conf/eacl/AbhishekAA17}, and {(vi)} \textbf{NFGEC+LME}~\cite{DBLP:conf/emnlp/XinZH0S18}~\footnote{We used code shared by authors to compute results for BBN data.}. For all these models, we use the scores reported in the published papers, as they are computed using the same data settings as that of ours.

Note that our model is not comparable with the implementation of~\citeauthor{DBLP:conf/naacl/XuB18}~\shortcite{DBLP:conf/naacl/XuB18}, because \citeauthor{DBLP:conf/naacl/XuB18} 
changed the~\FGNET{} problem definition to single-label classification problem and updated the training and testing data accordingly. It 
is hard to transform their work for multi-label, multi-class classification settings.% as their model performance declines drastically for our settings.%multi-label classification.

\begin{table*}[htbp]
    \centering
    \resizebox{1.85\columnwidth}{!}{
        \begin{tabular}{l|c|ccc|ccc|ccc}
            \hline
            \multirow{2}{*}{\textbf{Adjacency Matrix}} &  \multirow{2}{*}{\textbf{Model}} &  & Wiki &  &  & OntoNotes &  &  & BBN &  \\ 
            \cline{3-11}
            \multicolumn{1}{c|}{} & \multicolumn{1}{l|}{} & \multicolumn{1}{c}{strict} & \multicolumn{1}{c}{mac-F1} & \multicolumn{1}{c|}{mic-F1} & \multicolumn{1}{c}{strict} & \multicolumn{1}{c}{mac-F1} & \multicolumn{1}{c|}{mic-F1} & \multicolumn{1}{c}{strict} & \multicolumn{1}{c}{mac-F1} & \multicolumn{1}{c}{mic-F1} \\ \hline \hline
            
            \multicolumn{1}{c|}{} & \multicolumn{1}{l|}{\emph{mention + context}} & \multicolumn{1}{c}{0.649} & \multicolumn{1}{c}{0.802} & \multicolumn{1}{c|}{0.762} & \multicolumn{1}{c}{0.521} & \multicolumn{1}{c}{0.690} & \multicolumn{1}{c|}{0.632} & \multicolumn{1}{c}{0.612} & \multicolumn{1}{c}{0.746} & \multicolumn{1}{c}{0.759} \\
            
            \multicolumn{1}{l|}{Phase-I } & \multicolumn{1}{l|}{\emph{mention + context + position}} & \multicolumn{1}{c}{0.661} & \multicolumn{1}{c}{0.807} & \multicolumn{1}{c|}{0.767} & \multicolumn{1}{c}{0.531} & \multicolumn{1}{c}{0.694} & \multicolumn{1}{c|}{0.638} & \multicolumn{1}{c}{0.616} & \multicolumn{1}{c}{0.755} & \multicolumn{1}{c}{0.765} \\
            \hline
            
            \multicolumn{1}{l|}{Phase I-II (\FGGCN{} + {RND})} & \multicolumn{1}{l|}{\emph{mention + context + position + GCN}} & \multicolumn{1}{c}{0.642} & \multicolumn{1}{c}{0.797} & \multicolumn{1}{c|}{0.755} & \multicolumn{1}{c}{0.464} & \multicolumn{1}{c}{0.641} & \multicolumn{1}{c|}{0.595} & \multicolumn{1}{c}{0.617} & \multicolumn{1}{c}{0.693} & \multicolumn{1}{c}{0.709} \\
            
            \multicolumn{1}{l|}{Phase I-II (\FGGCN{} + {EYE})} & \multicolumn{1}{l|}{\emph{mention + context + position + GCN}} & \multicolumn{1}{c}{0.664} & \multicolumn{1}{c}{0.812} & \multicolumn{1}{c|}{0.773} & \multicolumn{1}{c}{0.519} & \multicolumn{1}{c}{0.681} & \multicolumn{1}{c|}{0.630} & \multicolumn{1}{c}{0.659} & \multicolumn{1}{c}{0.754} & \multicolumn{1}{c}{0.766} \\
            
            \multicolumn{1}{l|}{Phase I-II (\FGGCN{} + {PIVOTS})} & \multicolumn{1}{l|}{\emph{mention + context + position + GCN}} & \multicolumn{1}{c}{{0.672}} & \multicolumn{1}{c}{{0.815}} & \multicolumn{1}{c|}{{0.775}} & \multicolumn{1}{c}{\textbf{0.570}} & \multicolumn{1}{c}{{0.735}} & \multicolumn{1}{c|}{{0.675}} & \multicolumn{1}{c}{{0.736}} & \multicolumn{1}{c}{{0.804}} & \multicolumn{1}{c}{{0.810}} \\
            
            %\multicolumn{1}{l|}{Phase I-II (glove + \textit{pivots})} & \multicolumn{1}{l|}{\emph{mention + context + position + GCN}} & \multicolumn{1}{c}{{0.-}} & \multicolumn{1}{c}{{0.-}} & \multicolumn{1}{c|}{{0.-}} & \multicolumn{1}{c}{\textbf{0.572}} & \multicolumn{1}{c}{\textbf{0.736}} & \multicolumn{1}{c|}{\textbf{0.677}} & \multicolumn{1}{c}{{0.-}} & \multicolumn{1}{c}{{0.-}} & \multicolumn{1}{c}{{0.-}} \\ 
            
            \multicolumn{1}{l|}{Phase I-II (\FGGCN{} + {ATTN})} & \multicolumn{1}{l|}{\emph{mention + context + position + GCN}} & \multicolumn{1}{c}{\textbf{0.674}} & \multicolumn{1}{c}{\textbf{0.817}} & \multicolumn{1}{c|}{\textbf{0.777}} & \multicolumn{1}{c}{{0.567}} & \multicolumn{1}{c}{\textbf{0.737}} & \multicolumn{1}{c|}{\textbf{0.680}} & \multicolumn{1}{c}{\textbf{0.740}} & \multicolumn{1}{c}{\textbf{0.811}} & \multicolumn{1}{c}{\textbf{0.817}} \\ \hline
            
    \end{tabular}}
    \caption{Ablation study for \FGGCN{} using Glove + Context Encoder}
    \label{tab:Feature_Analysis}
\end{table*}

\subsection{Main Results}

We compare the results of our proposed approach (\FGGCN{}) with the baseline models 
in Table~\ref{tab:Results}. We boldface the overall best scores with the previous state-of-the-art underlined.
%The results in Table~\ref{tab:Results} show that our proposed model (\FGGCN{}) 
These results show that \FGGCN{} outperforms all the previous state-of-the-art 
research by a large margin. Especially noteworthy is the performance of our model 
on the BBN data outclassing the existing models by a margin of 10.4\%, 10.2\% and 8.3\% in 
strict accuracy, macro-f1, and micro-f1 respectively. For OntoNotes, our model yields 
5.1\% improvement in micro-f1 compared to the previous best by~\citeauthor{DBLP:conf/emnlp/XinZH0S18}~\shortcite{DBLP:conf/emnlp/XinZH0S18}. 
For Wiki data, the \FGGCN{} improves the performance by 7.5\%, 4.3\% and 4.0\% in strict accuracy, macro-f1, and micro-f1 respectively.
Such promising results show that refining the mention representations via corpus-level contextual clues help in alleviating the label noise associated with the distantly 
supervised data.

\subsection{Ablation study}
We present a detailed ablation analysis of the \FGGCN{} in Table~\ref{tab:Feature_Analysis}. Note that we only report the results for the Glove embeddings along with the context encoders. A similar trend is observed by replacing the Glove embeddings and contextual encoders with the deep contextualized ELMO embeddings.

%To evaluate the contribution of the individual components of \FGGCN{}, we conducted a detailed ablation study. Corresponding results are shown in Table~\ref{tab:Feature_Analysis}. 

For Phase-I, we analyze the role of position encoder in addition to the mention and the context encoders. We also compare the results of our model with noisy mention representations (Phase-I) with that of refined representations (Phase I-II). For Phase I-II, we examine the impact of variants of the adjacency matrix on representations' refinement, namely: {(i)} random adjacency matrix (\FGGCN{} + RND);
{(ii)} identity as the adjacency matrix (\FGGCN{} + EYE); 
{(iii)} adjacency matrix based on pivots, {i.e.,} $\eta_{ij} = 1$ (\FGGCN{} + PIVOTS); and
{(iv)} edge-weighed attention along with the adjacency matrix (\FGGCN{} + ATTN).

First two rows in Table~\ref{tab:Feature_Analysis} show that the \emph{position} feature slightly 
improved the performance for all data sets, yielding higher scores across all categories.
%We also report the performance of our model with randomly generated adjacency matrix 
% in order to analyze the the performance gained by the pivots.
Comparing the results for different adjacency matrices, we observe that the identity matrix didn't play a significant role in improving the model performance. For randomly generated adjacency matrices, a decline in performance shows that when the structure in data is lost the graph convolution layers are no longer useful 
in de-noising the representations. Note that we use randomly generated adjacency matrices with an 
equivalent number of edges as that of the original graphs (Section~\ref{experimental_settings}).

On the contrary, the models with adjacency matrices acquired from type-specific pivots, (\FGGCN{} + PIVOTS; \FGGCN{} + ATTN) show a substantial improvement in the performance, which is governed by an appropriate refinement of the noisy representations trained on distantly supervised training data. Especially, the edge-weighted attention yields a much higher score, because attention helps the model to reinforce the decision by attending over the contribution of each neighbor. Overall, the results show that the refined representations indeed augment the model performance by a significant margin. %accommodating the corpus-level contextual clues.% from a global perspective. 

\subsection{Analyses}
\label{all:analysis}
In this section, we analyze the effectiveness of the refined representations (Phase-II), followed by a detailed 
analysis of the error cases.

\paragraph{Effectiveness of Phase-II (BBN data):}
In order to analyze the effectiveness of the refined representations (Phase-II), we perform a comparative performance analysis for the most frequent entity types in the BBN dataset. %We choose BBN dataset for analysis because it has a relatively large proportion of testing mentions.

As shown in Table~\ref{tab:phase_analysis}, Phase-II has a clear dominance over Phase-I across all entity types.
The major reason for poor performance in Phase-I is highly correlated nature of entity types with substantial 
contextual overlap. The distant supervision, moreover, adds to the intricacy of the problem.
For example, \emph{``organization"} and \emph{``cooperation"} are two highly correlated entity types, 
\emph{``organization"} is a generic term, whereas, the \emph{``corporation"} being a sub-type, is 
more intended towards an enterprise and/or company with some business-related concerns, {i.e.,} 
every corporation is an organization but not otherwise.
Analyzing the corpus-level additive lexical contrast along the type-hierarchy revealed that 
in addition to sharing the context of the \emph{``organization"}, the context of the \emph{``corporation"} is 
more oriented towards tokens: \emph{cents}, \emph{business}, \emph{stake}, \emph{bank}, {etc.} 
%Likewise, the entity type \emph{``government organization"} adds the following distinctive tokens: \emph{budget}, \emph{fiscal}, \emph{deficit}, \emph{bills}, {etc.,} to the context of \emph{``organization"}. 
However, for distantly supervised data, it is hard to ensure such a distinctive distribution of contextual 
tokens for each entity mention. In addition, the lack of information sharing among these distinctive tokens 
across sentence boundaries leads to poor performance for Phase-I, which makes it a much harder problem from the
generalization perspective. A similar scenario holds for other entity types, {e.g.,} \emph{actor} vs \emph{artist} vs \emph{director}, {etc.} 
%For such tricky cases, the mention representations solely based on local context are inadequate for \FGNET{}. 
Whereas, after sharing type-specific contextual clues, we can see a drastic improvement in the performance for 
phase I-II, i.e., F1 = 0.844 compared to F1 = 0.790 in phase-I for \emph{``corporation"}, shown in Table~\ref{tab:phase_analysis}.

To further verify our claim, we analyze the nearest neighbors for both the noisy and the refined 
representation spaces along with the corresponding labels. An example in this regard is shown in 
Table~\ref{tab:nearest_neighbors}, where we illustrate the neighboring representations corresponding 
to the representation of the mention \emph{``Maytag"} from the sentence: \emph{``We have officials from giants like 
    Du Pont and \textbf{Maytag}, along with lesser knowns like Trojan Steel and the Valley Queen 
    Cheese Factory."} with true context-dependent label as \emph{``organization/corporation"}.
\begin{table}[]
    \resizebox{.99\columnwidth}{!}{
    \begin{tabular}{l|c|c c c|c c c}
        \hline
        \multirow{2}{*}{Labels}    & \multirow{2}{*}{Support} &       & Phase-I &       &       & Phase I-II &       \\
        &                          & Prec  & Rec     & F1    & Prec  & Rec        & F1    \\
        \hline
        /organization             & 45.30\%                  & 0.837 & 0.850   & 0.843 & 0.924 & 0.842      & \textbf{0.881} \\
        /organization/corporation & 35.70\%                  & 0.824 & 0.759   & 0.790  & 0.921 & 0.779      & \textbf{0.844} \\
        /person                   & 22.00\%                  & 0.746 & 0.779   & 0.762 & 0.86  & 0.886      & \textbf{0.872} \\
        /gpe                      & 21.30\%                  & 0.878 & 0.831   & 0.853 & 0.924 & 0.845      & \textbf{0.883} \\
        /gpe/city                 & 9.17\%                   & 0.737 & 0.738   & 0.738 & 0.802 & 0.767      & \textbf{0.784} \\
        %/gpe/country              & 8.50\%                   & 0.952 & 0.465   & 0.625 & 0.941 & 0.743      & \textbf{0.830}  \\
        %/organization/government  & 6.10\%                   & 0.765 & 0.512   & 0.613 & 0.739 & 0.674      & \textbf{0.705}\\
        \hline
    \end{tabular}}
    \caption{Phase-I vs Phase (I-II) comparison for BBN data}
    \label{tab:phase_analysis}
\end{table}
The neighboring representations corresponding to the noisy representation space are dominated by irrelevant 
labels, having almost no correlation with the original mention's label. Whereas, for the refined representations, 
almost all the neighboring representations carry the same label as that of mention \emph{``Maytag"}.
This ascertains that the refined representations are more semantically oriented \emph{w.r.t} context, which enables 
them to accommodate more distinguishing information for entity typing compared to that of the noisy representations.

These analyses strengthen our claim that FGET-RR enables representation refinement and/or label smoothing by 
implicitly sharing corpus-level contextual clues across entity mentions. This empowers FGET-RR to indeed learn 
across sentence boundaries, which makes it more robust compared with the previous state-of-the-art methods 
that classify entity mentions entirely independent of each other.

% \usepackage{graphicx}
\begin{table}[]
    \resizebox{0.99\columnwidth}{!}{%
        \begin{tabular}{ll|ll}
            \hline
            \multicolumn{2}{l|}{Noisy Representations (Phase-I)} & \multicolumn{2}{l}{Refined Representations (Phase I-II)} \\
            %\hline
            Mention & Label & Mention & Label \\
            \hline
            Yves Goupil      & /person                          & Ford Motor                  & /organization/corporation \\
            Berg             & /person                          & Vauxhall Motors Ltd.        & /organization/corporation  \\
            Volokhs          & /person                          & Chrysler Corp.              & /organization/corporation  \\
            lawns            & /plant                           & Advanced Micro Devices Inc. & /organization/corporation  \\
            Rafales          & /product/vehicle,/product      & Chesebrough-Pond s Inc.'    & /organization/corporation  \\
            %Katonah          & /gpe/city, '/gpe'                & Diamond-Star Motors Corp.   & /organization/corporation \\
            \hline 
        \end{tabular}%
    }
    \caption{Top 5-nearest neighboring representations (noisy and refined) for the representation of mention\emph{``Maytag"}}
    \label{tab:nearest_neighbors}
\end{table}

\paragraph{Error Cases:}

We categorize the errors into two major categories: {(i)} missing labels, and {(ii)} erroneous labels. 
Missing labels correspond to the entity mentions for which type labels are not predicted by the model, 
thus effecting the recall, while erroneous labels correspond to mis-labeled instances, effecting the 
precision. Following the results for the BBN data in Table~\ref{tab:phase_analysis}, most of the errors 
(for phase-I and I-II) correspond to the labels \emph{"/organization/corporation"} and \emph{``organization"}.

For missing labels, most of them are attributed to the cases, where type labels are entirely dictated by the names of corporations, with very little information contained in the context. For example, in 
the sentence: \emph{``That has got to cause people feel a little more optimistic, says 
    Glenn Cox the correspondence officer of Mcjunkin"}, the entity mention \emph{``Mcjunkin"} is labeled 
\emph{``organization/corporation"}. For such cases, type information is not explicit from the context. 
This is also evident by a relatively low recall score for both Phase-I and Phase I-II, shown in Table~\ref{tab:phase_analysis}.

Likewise, most of the erroneous labels (esp., Phase-I) are caused by the overlapping context for highly 
correlated entity types, {e.g.,} \emph{``organization"} and \emph{``corporation"}, as explained previously. 
This problem was somehow eradicated by refining the representations in Phase-II, as is evident by a higher 
change in precision for Phase I-II relative to that of Phase-I. A similar trend was observed for OntoNotes and Wiki data. 
%For OntoNotes, a relatively higher proportion of errors is caused by the pronominal mentions. 
Other limiting factors of the proposed model include: {(i)} the assumption that ELMO embeddings are able to capture distinctive mention representation based on the context, {(ii)} acquiring pivot vectors from noisy data, which did some smoothing but didn't completely rectify the noise.

%%% Conclusion
%\input{5_Conclusions.tex}

\section{Conclusions and Future Work}
\label{conclusions_}

In this paper, we propose \FGGCN{}, a novel approach for \FGNET{} that outperforms existing research by a large margin. In the future, we will augment the proposed framework by explicitly identifying type-specific clauses to perform edge conditioned representations' refinement.

%%% Acknowledgements
%\paragraph{Acknowledgments.}
%This research was partially funded by ARC DPs 170103710 and 180103411, and D2DCRC DC25002 and DC25003.

% Concluding Lines

\paragraph*{Acknowledgements.}%
This work is supported by ARC DPs 170103710 and 180103411, D2DCRC DC25002 and DC25003. The Titan V used for this research was donated by the NVIDIA Corporation.

\bibliography{aaai2020}
\bibliographystyle{aaai} 
\end{document}